\documentclass[conference]{IEEEtran}

\usepackage{cite}
\usepackage{amsmath,amssymb,amsfonts}
\usepackage{algorithmic}
\usepackage{graphicx}
\usepackage{textcomp}
\usepackage{xcolor}
\usepackage{booktabs}
\usepackage{multirow}
\usepackage{hyperref}
\usepackage{array}
\usepackage{tabularx}
\usepackage{subcaption}

\def\BibTeX{{\rm B\kern-.05em{\sc i\kern-.025em b}\kern-.08em
    T\kern-.1667em\lower.7ex\hbox{E}\kern-.125emX}}

\begin{document}

\title{Graph-Augmented Retrieval for Cross-Entity Financial Sentiment Analysis: A Comparative Study}

\author{
\IEEEauthorblockN{Rajan Bastakoti}
\IEEEauthorblockA{
DePaul University\\
Chicago, IL, USA\\
r.bastakoti@depaul.edu}
\and
\IEEEauthorblockN{Sagar Bhetwal}
\IEEEauthorblockA{
Youngstwon State University\\
Youngstown, Ohio, USA\\
ragasvhetwal9841@gmail.com}
\and
\IEEEauthorblockN{Nirajan Acharya}
\IEEEauthorblockA{
Youngstown State University\\
Youngstown, Ohio, USA\\
nirajanach3@gmail.com}
\and
\IEEEauthorblockN{Gaurav Kumar Gupta}
\IEEEauthorblockA{
Youngstown State University\\
Youngstown, Ohio, USA\\
guptagauravk1@gmail.com}
}

\maketitle

\begin{abstract}
Retrieval-Augmented Generation (RAG) has become foundational for grounding large language models in domain-specific corpora, yet conventional vector-based RAG systems are fundamentally limited in their ability to capture the structured, multi-entity relationships that underpin financial market analysis. This paper presents a comprehensive comparative study of a novel two-hop Graph-RAG architecture versus a standard vector-only baseline for cross-entity financial sentiment analysis. Our system constructs a sentiment-weighted knowledge graph of 59 equity entities from 255 news articles covering 10 major technology stocks, then augments dense retrieval with intensity-filtered graph traversal over \texttt{INFLUENCES} edges to surface relational evidence inaccessible to vector search alone. We evaluate both architectures on 100 grounded queries (30 Direct, 70 Relational) using semantic similarity, entity recall, RAGAS metrics, latency benchmarks, and ablation studies. Graph-RAG achieves a statistically significant improvement in entity recall ($+6.4\%$, $p < 0.001$, Wilcoxon signed-rank) and delivers substantially more relevant answers for complex multi-entity queries ($+11.7\%$ Answer Relevancy), with gains concentrating in relational question types ($+16.1\%$). Critically, these improvements come at no measurable cost to answer quality ($\Delta = +0.001$ semantic similarity, Cohen's $d = 0.078$), with a modest $22.6\%$ increase in mean latency offset by an $80\%$ reduction in latency variance. An ablation study on the graph traversal intensity threshold reveals an inverted-U relationship with answer quality, identifying $\tau = 0.5$ as optimal over the production default of $\tau = 0.7$. These findings characterize a precision-for-coverage trade-off inherent to graph-augmented retrieval and provide actionable architectural guidance for practitioners building RAG systems for multi-entity financial analysis.
\end{abstract}


\section{Introduction}

The rapid digitization of global financial markets has resulted in an unprecedented volume of unstructured textual data, making real-time equity synthesis a critical challenge for financial analysts. To navigate this complexity, Large Language Models (LLMs) have been increasingly deployed within Retrieval-Augmented Generation (RAG) frameworks to ground generative outputs in authoritative external corpora \cite{lewis2020rag}. However, as the domain shifts toward specialized financial market analysis, traditional ``vector-only'' RAG architectures are encountering a significant ``contextual plateau.'' While effective for localized fact-retrieval, these systems remain fundamentally blind to the structured interdependencies---such as supply-chain cascades and multi-hop influence networks---that define systemic market risk \cite{barry2025graphrag}.

Standard vector-based retrieval operates on the assumption that semantic proximity in a latent vector space is a sufficient proxy for informational relevance. In the financial sector, this paradigm faces three critical technical limitations. First, traditional RAG treats text as isolated chunks, neglecting the explicit structured relationships required to navigate multi-hop logic \cite{barry2025graphrag}. This results in an inability to perform ``global sensemaking'' or query-focused summarization across a corpus to identify high-level market trends \cite{edge2024graphrag}. Second, the concatenation of multiple semantically similar text snippets often leads to the ``lost in the middle'' phenomenon, where crucial financial details are obscured within redundant, high-token-count contexts \cite{barry2025graphrag, es2023ragas}. Finally, financial terminology is highly susceptible to ``semantic noise''; without a structured graph to logically separate entities, semantic overlap in dense retrieval can lead to critical factual errors in analysis \cite{chen2024knowledge}.

Furthermore, recent research into intertextual connections has highlighted the importance of sentiment ``contagion''---the propagation of sentiment across documents and entities \cite{nasiopoulos2025financial}. While purely lexical analysis often overlooks the discursive signals of how sentiment spreads from one market participant to another, a Graph-Augmented approach allows for the modeling of these ``ripples.'' Unlike ``Hybrid RAG,'' which often switches between keyword and vector search, a true ``Graph-RAG'' architecture traverses a pre-constructed knowledge graph, using retrieved chunks as ``seeds'' to explore adjacent entities and relationship triples \cite{barry2025graphrag, edge2024graphrag}. This not only improves context precision but also significantly condenses context density---often requiring a 9x to 43x reduction in token consumption compared to traditional map-reduce approaches \cite{barry2025graphrag}.

To address these limitations, this paper proposes and evaluates a Graph-Augmented RAG framework utilizing a high-intensity network of 59 distinct equity entities. By mapping news vectors onto a multi-layered Neo4j knowledge graph, our architecture enables systemic traversals that reconcile narrative-driven news headlines with quantitative influence and sentiment intensity scores. We evaluate this framework against a baseline General RAG system using the RAGAS (Retrieval-Augmented Generation Assessment) framework, focusing on Faithfulness and Context Relevancy \cite{es2023ragas}. Our research demonstrates that by integrating a specialized 59-entity influence network, the system can effectively identify ``market dissonance''---instances where the textual narrative of news is corrected by the underlying metadata of the equity graph---providing a more robust risk profile for automated financial market analysis.

\section{Related Work}

The evolution of Retrieval-Augmented Generation (RAG) has been primarily driven by the need to ground Large Language Models (LLMs) in verifiable facts \cite{lewis2020rag}. However, as implementations move into specialized domains like finance, the limitations of traditional vector-based retrieval have become a critical focus of academic inquiry.

\subsection{The Semantic Search Bottleneck in Traditional RAG}

Recent research identifies a fundamental ``bottleneck'' in standard vector RAG when tasked with complex, cross-document reasoning. Edge et al. \cite{edge2024graphrag} highlight that conventional dense semantic retrieval is optimized for finding localized, explicitly stated facts but fails on ``sensemaking'' queries that require a global understanding of an entire dataset. Barry et al. \cite{barry2025graphrag} argue that because vector RAG treats text as isolated chunks, it inherently neglects the explicit structured relationships---such as supply chains or regulatory dependencies---critical for multi-hop logic. Furthermore, when tasked with cross-document comparisons, vector-only systems suffer from severe computational complexity, often requiring expensive pairwise comparisons that result in redundant, excessively lengthy context windows. This leads to the ``lost in the middle'' phenomenon, where LLMs lose focus on crucial details buried within long, concatenated passages \cite{barry2025graphrag, es2023ragas}.

\subsection{Structural Knowledge in Financial Informatics}

The justification for utilizing structured Knowledge Graphs (KG) over raw text is particularly strong in financial report generation. Chen et al. \cite{chen2024knowledge} note that expert-level financial data is often sparsely distributed across free-form text, making it highly susceptible to ``semantic noise.'' Structured knowledge allows for the filtering of this noise by explicitly mapping sequential, causal, and hypernym-hyponym relations between entities \cite{chen2024knowledge}. Zehra et al. \cite{zehra2021fkg} further emphasize that annual reports lack standardization in format and vocabulary, which hinders automated extraction from raw text. In contrast, a Financial Knowledge Graph (FKG) provides a logical map that reduces redundancy and token overhead. Additionally, Nasiopoulos et al. \cite{nasiopoulos2025financial} introduce the concept of sentiment ``contagion,'' demonstrating that analyzing intertextual connections considerably improves predictive accuracy compared to purely lexical analysis. This underscores the necessity of a networked approach to capture how sentiment cascades across an equity ecosystem.

\subsection{Architectural Paradigms: Implicit vs.\ Explicit Retrieval}

The literature distinguishes between ``Implicit'' (sub-symbolic) and ``Explicit'' (symbolic) retrieval within modern Hybrid RAG architectures. Implicit retrieval utilizes conventional vector embeddings to find $k$-nearest neighbors, while Explicit retrieval translates natural language into structured queries, such as text-to-Cypher, to navigate a KG \cite{barry2025graphrag}. To bridge these methods, researchers have proposed the use of ``seed'' nodes. In this logic, traditional semantic search identifies a subset of relevant text chunks which then act as anchors to explore the graph, retrieving adjacent entities and triples that a vector search alone would miss \cite{barry2025graphrag}. For global summarization, Edge et al. \cite{edge2024graphrag} propose partitioning the KG into a hierarchy of ``communities.'' By pre-generating summaries for these modular groups, the system can perform a map-reduce process to aggregate a holistic response from an entire corpus, significantly reducing token consumption.

\subsection{Reference-Free Evaluation Frameworks}

A significant challenge in RAG development is the lack of human-annotated ground-truth datasets for real-time evaluation. Es et al. \cite{es2023ragas} argue that traditional NLP metrics like BLEU or ROUGE are insufficient, as they rely on reference answers and often fail to predict downstream performance in long-form generation. To address this, the RAGAS framework advocates for LLM-based, ``reference-free'' metrics. By utilizing an LLM to judge specific dimensions---Faithfulness (groundedness), Context Relevance (signal-to-noise ratio), and Answer Relevance---developers can estimate system correctness without the bottleneck of human annotation, allowing for faster iterative cycles in RAG architecture design \cite{barry2025graphrag, es2023ragas}.

\section{Methodology}

\begin{figure}[htbp]
\centerline{\includegraphics[width=\columnwidth]{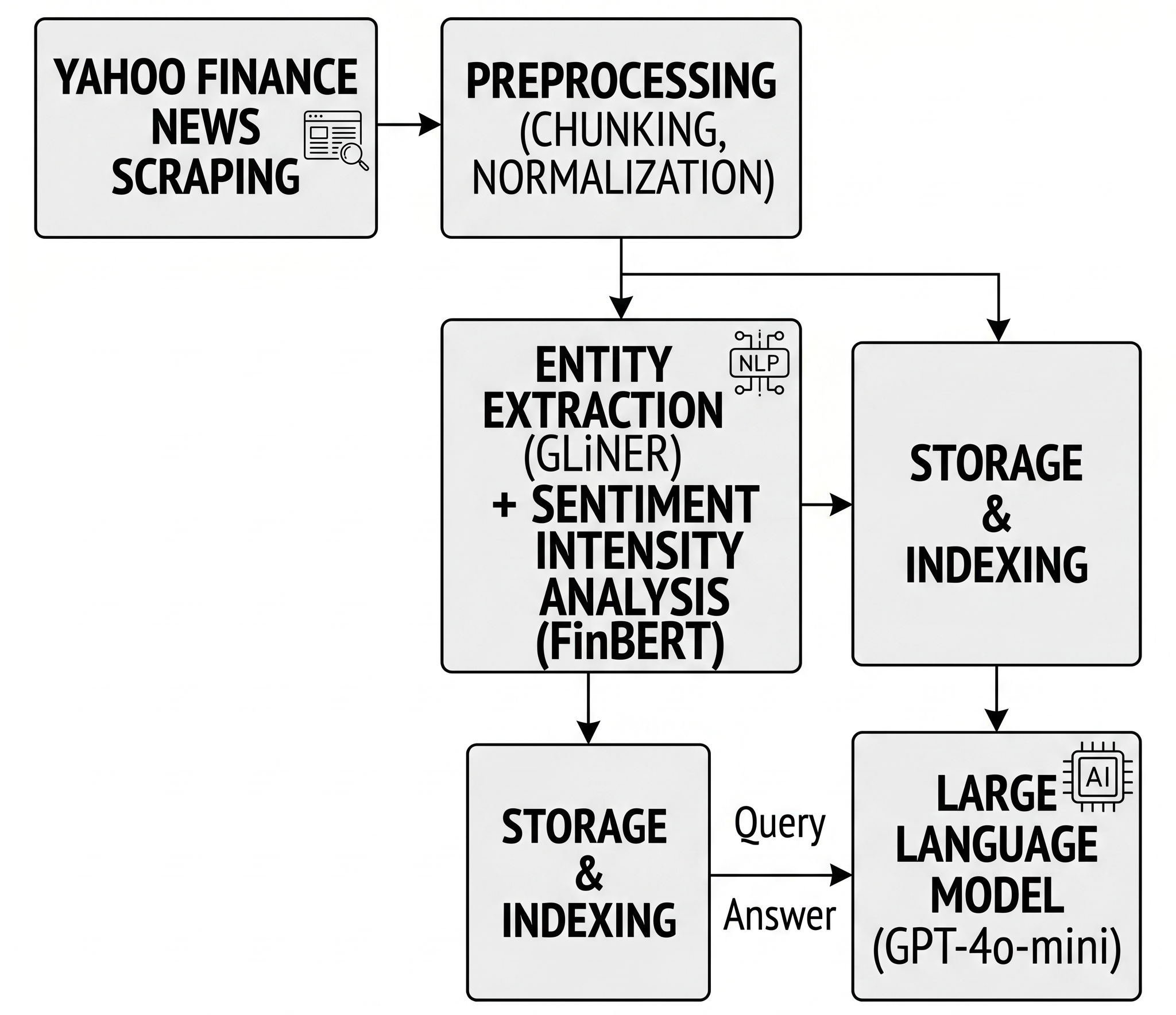}}
\caption{Overall system architecture for Graph-RAG financial synthesis. The pipeline spans from automated corpus construction through knowledge graph ingestion to dual-path retrieval and LLM generation.}
\label{fig:system_arch}
\end{figure}

The research utilizes a dual-architecture approach to quantify the impact of graph-based relational context on financial synthesis. The methodology transitions from a high-concurrency data ingestion phase to a comparative retrieval study between a vector-only baseline and the proposed 2-hop graph framework. The overall system architecture is illustrated in Fig.~\ref{fig:system_arch}.

\subsection{Automated Corpus Construction and Pre-processing}

The news corpus was generated using a high-concurrency extraction engine targeting ten high-liquidity primary tickers ($T_{primary} = \{$AAPL, GOOGL, MSFT, AMZN, TSLA, NVDA, META, NFLX, PLTR, DIS$\}$). Using Playwright and BeautifulSoup, the engine performed iterative browser scrolling to capture real-time disclosures, resulting in 300+ articles with associated metadata (URLs and publication dates).

The raw text was fragmented into 1,477 distinct news chunks. Each chunk was encoded into a 1,024-dimensional latent space using the BAAI/bge-large-en-v1.5 transformer model. This model was selected for its state-of-the-art performance in retrieval tasks on the Massive Text Embedding Benchmark (MTEB).

\subsection{Relational Impact Modeling and Graph Topology}

To build the knowledge graph, unstructured text was transformed into structured triples through a dual-model pipeline:

\begin{enumerate}
    \item \textbf{Entity Extraction (GLiNER):} We utilized the knowledgator/gliner-bi-large-v2.0 model for zero-shot recognition of custom labels: ``Stock Ticker,'' ``Equity Index,'' ``Investment Bank,'' and ``Organization.''
    \item \textbf{Sentiment Weighting (FinBERT):} Sentiment polarity was calculated via ProsusAI/finbert. The relational Intensity ($I$) between entities was defined as the absolute difference between positive and negative scores:
    \begin{equation}
        I = |Score_{pos} - Score_{neg}|
    \end{equation}
\end{enumerate}

\begin{figure}[htbp]
\centerline{\includegraphics[width=\columnwidth]{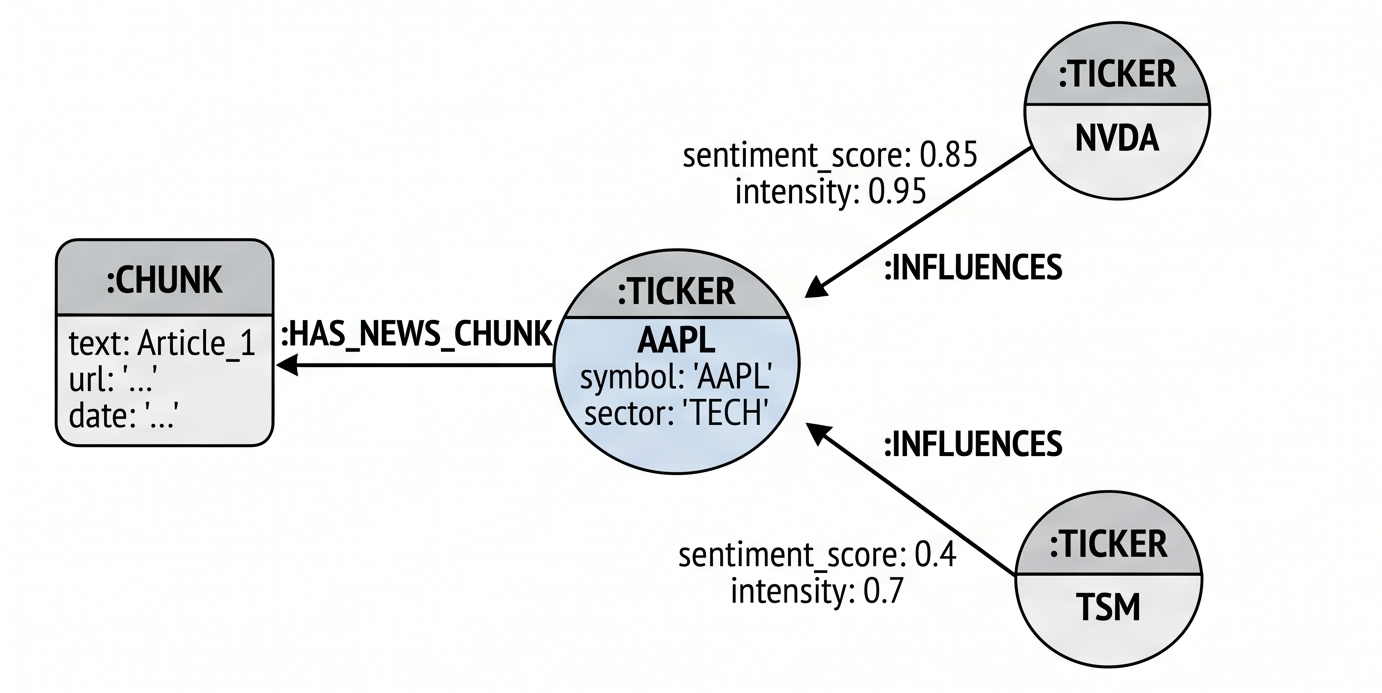}}
\caption{Knowledge graph schema for equity entities influence network. Ticker nodes are connected via sentiment-weighted INFLUENCES edges, with news chunks attached as evidence.}
\label{fig:kg_schema}
\end{figure}

The normalized entities resulted in a Neo4j graph topology containing 59 canonical equity entities. The schema maps \texttt{Ticker} nodes to \texttt{Chunk} nodes via \texttt{HAS\_NEWS\_CHUNK} edges, while inter-ticker relationships are defined by directed \texttt{INFLUENCES} edges weighted by sentiment and intensity. The knowledge graph schema is shown in Fig.~\ref{fig:kg_schema}.

\subsection{Baseline Architecture: General RAG (Vector-Only)}

The General RAG system serves as the control baseline, representing a standard industry implementation of dense semantic retrieval.

\begin{itemize}
    \item \textbf{Indexing:} All 1,477 chunk embeddings were indexed using FAISS (Facebook AI Similarity Search) with an \texttt{IndexFlatIP} (Inner Product) configuration for rapid exact $k$-nearest neighbor search.
    \item \textbf{Retrieval:} For a given query $q$, the system retrieves the top $k=5$ chunks based on maximum cosine similarity in the latent vector space.
    \item \textbf{Contextualization:} The retrieved segments are concatenated into a flat context window: ``--- SOURCE ARTICLE --- TEXT: \{chunk\_text\}''. No relational metadata or secondary entity information is included in this baseline.
\end{itemize}

\subsection{Proposed Architecture: 2-Hop Graph-RAG}

\begin{figure}[htbp]
\centerline{\includegraphics[width=\columnwidth]{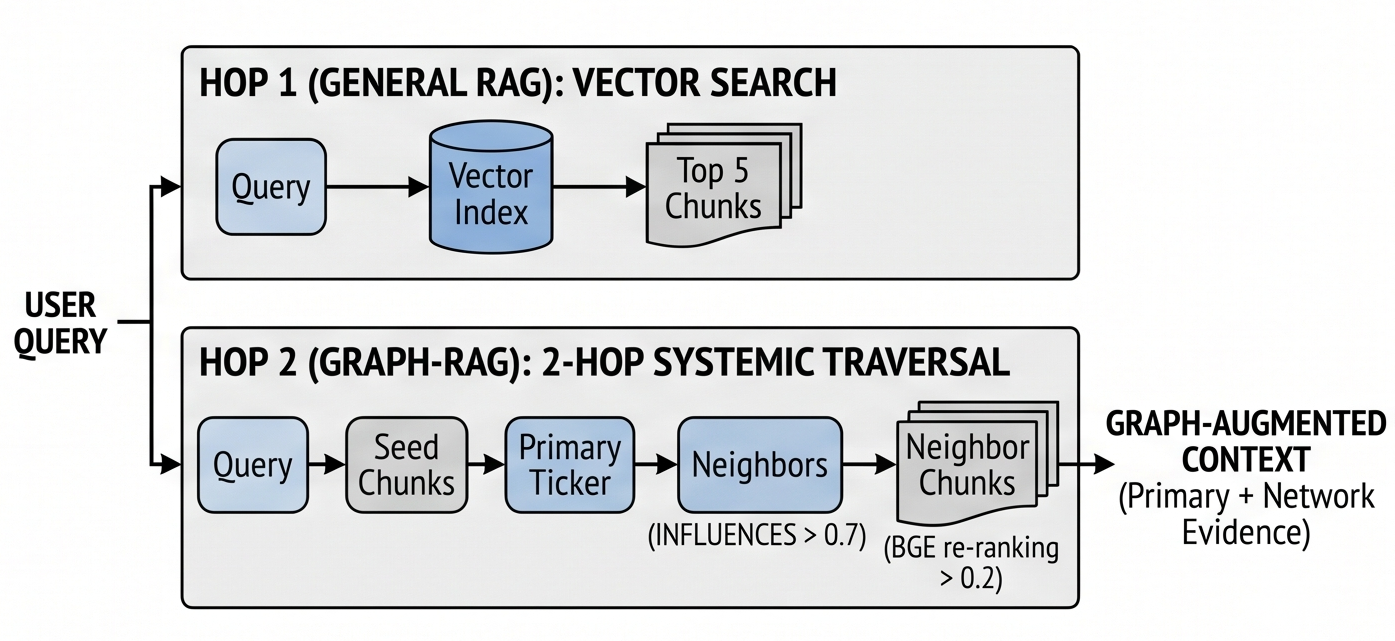}}
\caption{General RAG vs.\ Graph RAG logic diagram. The baseline retrieves only semantically similar chunks, while Graph-RAG performs structural expansion via intensity-filtered graph traversal.}
\label{fig:rag_comparison}
\end{figure}

The Graph-RAG framework leverages the Neo4j structure to perform systemic traversals that capture ``market ripples.'' The comparative logic between both architectures is illustrated in Fig.~\ref{fig:rag_comparison}.

\begin{itemize}
    \item \textbf{Hop 1 (Semantic Anchor):} The system performs a vector search within the Neo4j \texttt{news\_vectors} index to identify the top $k=5$ ``seed'' chunks.
    \item \textbf{Hop 2 (Structural Expansion):} From the ticker associated with each seed chunk, the system traverses \texttt{INFLUENCES} edges where $\text{Intensity} > 0.7$ to find neighboring entities.
    \item \textbf{Neighbor Re-ranking:} For each neighbor ticker, the system retrieves its associated news chunks and re-ranks them against the query using the BGE model. Only the top 2 most relevant chunks per neighbor with a similarity score $> 0.2$ are added as ``Network Evidence.''
    \item \textbf{Synthesis:} The LLM receives a graph-augmented context containing primary news, connected entity names, their numerical sentiment scores, and the re-ranked evidence chunks.
\end{itemize}

\subsection{Generation and Evaluation Framework}

Both systems utilize GPT-4o-mini (Azure OpenAI) for generation, employing identical parameters: a temperature of $0.2$ and a unified system prompt enforcing strict context grounding.

Performance was evaluated through a multi-tiered approach:

\begin{itemize}
    \item \textbf{Statistical Significance:} Semantic similarity (cosine similarity between answer and ground truth) and Entity Recall were calculated for the full dataset of 100 queries. Statistical significance was validated through Wilcoxon signed-rank tests performed on this comprehensive set.
    \item \textbf{RAGAS Metrics:} Due to high API consumption, the RAGAS framework \cite{es2023ragas} (Faithfulness, Answer Relevancy, Context Precision, and Context Recall) was performed on a stratified sample of 25 queries, ensuring proportional representation of ``Direct'' and ``Relational'' question types.
\end{itemize}

\section{Results}

This section presents the empirical findings from our comparative evaluation of the General RAG baseline and the proposed 2-hop Graph-RAG system across 100 evaluation queries.

\subsection{Overall Performance Comparison}

Table~\ref{tab:main_results} summarizes the aggregate performance of both retrieval architectures across all evaluation dimensions.

\begin{table}[htbp]
\caption{Aggregate Evaluation Results ($n=100$ queries)}
\label{tab:main_results}
\centering
\begin{tabular}{lcccc}
\toprule
\textbf{Metric} & \textbf{General} & \textbf{Graph} & \textbf{$\Delta$} & \textbf{Winner} \\
\midrule
Semantic Similarity & 0.8725 & 0.8737 & +0.001 & Tie \\
Entity Recall & 0.860 & 0.924 & +0.064 & \textbf{Graph}$^{***}$ \\
Signal Density$^\dagger$ & 2.696 & 1.689 & $-$37.4\% & General \\
Avg.\ Latency (s) & 7.37 & 9.03 & +22.6\% & General \\
Latency $\sigma$ (s) & 4.82 & 0.96 & $-$80.1\% & \textbf{Graph} \\
\midrule
Faithfulness & 0.8902 & 0.8702 & $-$0.020 & General \\
Answer Relevancy & 0.7512 & 0.8679 & +0.117 & \textbf{Graph} \\
Context Precision & 0.8689 & 0.9296 & +0.061 & \textbf{Graph} \\
Context Recall & 0.6486 & 0.6174 & $-$0.031 & General \\
\bottomrule
\multicolumn{5}{l}{\footnotesize $^{***}p < 0.001$ (Wilcoxon signed-rank test, one-sided)} \\
\multicolumn{5}{l}{\footnotesize $^\dagger$ Relevant entities per 1{,}000 context tokens} \\
\end{tabular}
\end{table}

The primary finding is that Graph-RAG achieves a statistically significant improvement in entity recall ($+6.4\%$, $p = 0.000043$, Wilcoxon signed-rank), confirming that structural graph traversal surfaces entities that pure vector similarity misses. Semantic similarity between the two systems is statistically indistinguishable ($\Delta = +0.001$, $p = 0.281$, Cohen's $d = 0.078$), confirming that the additional graph context neither improves nor degrades generation quality.

Among the RAGAS metrics~\cite{es2023ragas}, the most notable result is the $+11.7\%$ improvement in Answer Relevancy, suggesting that graph-augmented context helps the LLM produce more topically focused responses. Context Precision also favors Graph-RAG ($+6.1\%$), indicating that the re-ranking step effectively filters irrelevant neighbor evidence.

\subsection{Stratified Analysis by Question Type}

To isolate the effect of query complexity, we stratified the RAGAS evaluation into Direct (single-entity factual) and Relational (cross-entity impact) question types. Results are presented in Table~\ref{tab:stratified}.

\begin{table}[htbp]
\caption{RAGAS Metrics Stratified by Question Type}
\label{tab:stratified}
\centering
\begin{tabular}{llccc}
\toprule
\textbf{Type} & \textbf{Metric} & \textbf{General} & \textbf{Graph} & \textbf{$\Delta$} \\
\midrule
\multirow{4}{*}{Direct ($n=7$)} 
  & Faithfulness & 0.9841 & 0.9524 & $-$0.032 \\
  & Answer Relevancy & 0.8828 & 0.8909 & +0.008 \\
  & Context Precision & 0.8982 & 0.9839 & +0.086 \\
  & Context Recall & 0.8810 & 0.7619 & $-$0.119 \\
\midrule
\multirow{4}{*}{Relational ($n=17$)} 
  & Faithfulness & 0.8515 & 0.8363 & $-$0.015 \\
  & Answer Relevancy & 0.6971 & 0.8584 & \textbf{+0.161} \\
  & Context Precision & 0.8569 & 0.9072 & +0.050 \\
  & Context Recall & 0.5529 & 0.5578 & +0.005 \\
\bottomrule
\multicolumn{5}{l}{\footnotesize RAGAS evaluated on a 24-question stratified subsample.} \\
\end{tabular}
\end{table}

\begin{figure}[htbp]
\centerline{\includegraphics[width=\columnwidth]{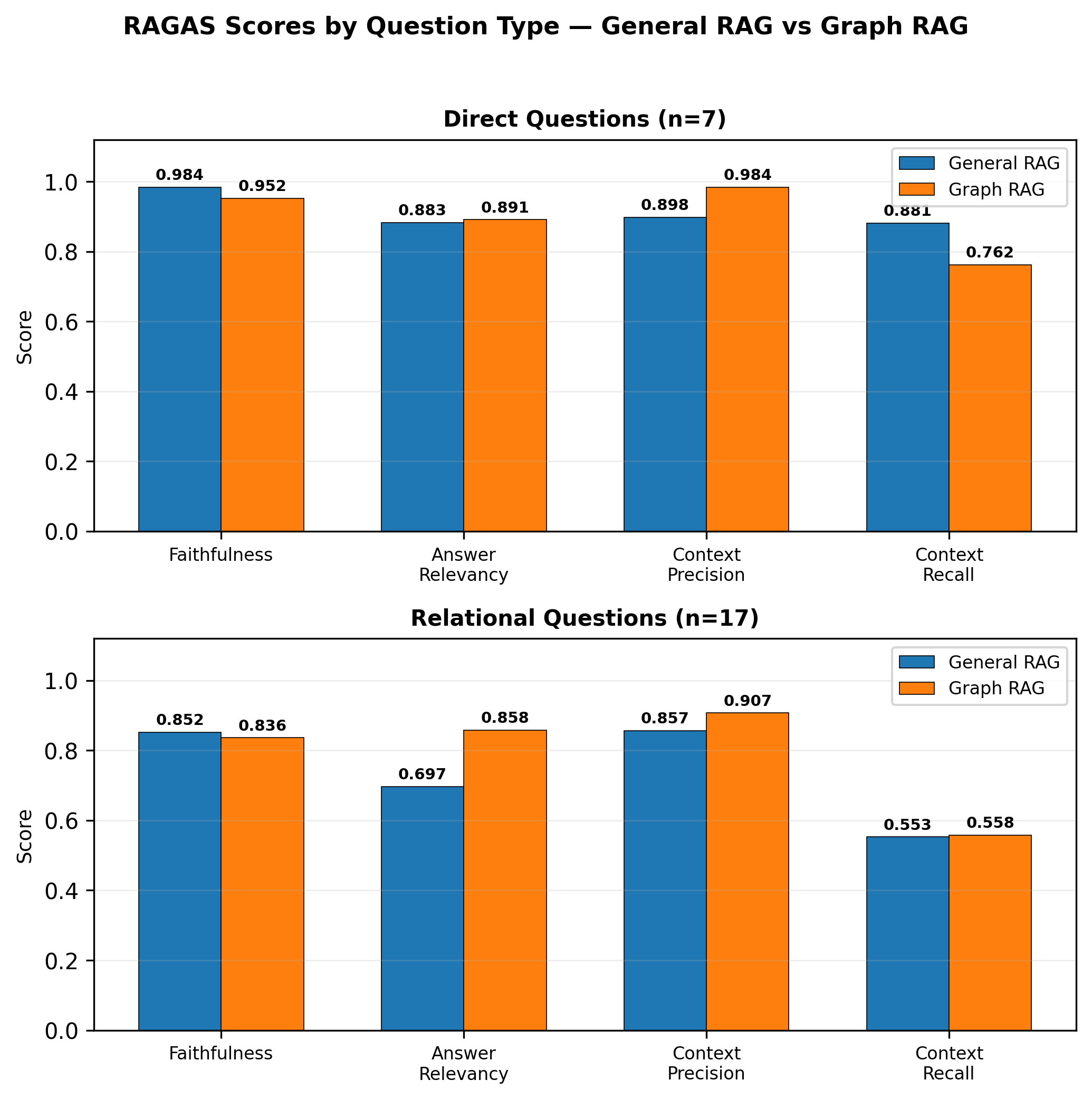}}
\caption{RAGAS scores stratified by question type. Graph-RAG's advantage is pronounced for Relational queries, particularly in Answer Relevancy ($+16.1\%$).}
\label{fig:ragas_qtype}
\end{figure}

Graph-RAG's advantage concentrates heavily in Relational queries, where Answer Relevancy improves by $+16.1\%$, as shown in Fig.~\ref{fig:ragas_qtype}. This is the strongest per-metric result in the study and demonstrates that the 2-hop expansion provides substantive value for multi-entity reasoning tasks. For Direct queries, both systems perform comparably, with Graph-RAG marginally improving Context Precision ($+8.6\%$) at the cost of lower Context Recall ($-11.9\%$)---an expected trade-off when additional entities dilute single-article evidence.

\subsection{Ablation Study: Intensity Threshold $\tau$}

The intensity threshold $\tau$ governs which \texttt{INFLUENCES} edges are traversed during the second hop. We evaluated three threshold values on a 20-question stratified subsample to characterize the sensitivity of the system. Results are shown in Table~\ref{tab:ablation}.

\begin{table}[htbp]
\caption{Ablation: Intensity Threshold vs.\ Semantic Similarity}
\label{tab:ablation}
\centering
\begin{tabular}{cc}
\toprule
\textbf{Threshold ($\tau$)} & \textbf{Mean Similarity} \\
\midrule
0.3 & 0.8388 \\
0.5 & \textbf{0.8550} \\
0.7 & 0.8501 \\
\bottomrule
\multicolumn{2}{l}{\footnotesize Evaluated on 20-question subsample.} \\
\end{tabular}
\end{table}

\begin{figure}[htbp]
\centerline{\includegraphics[width=\columnwidth]{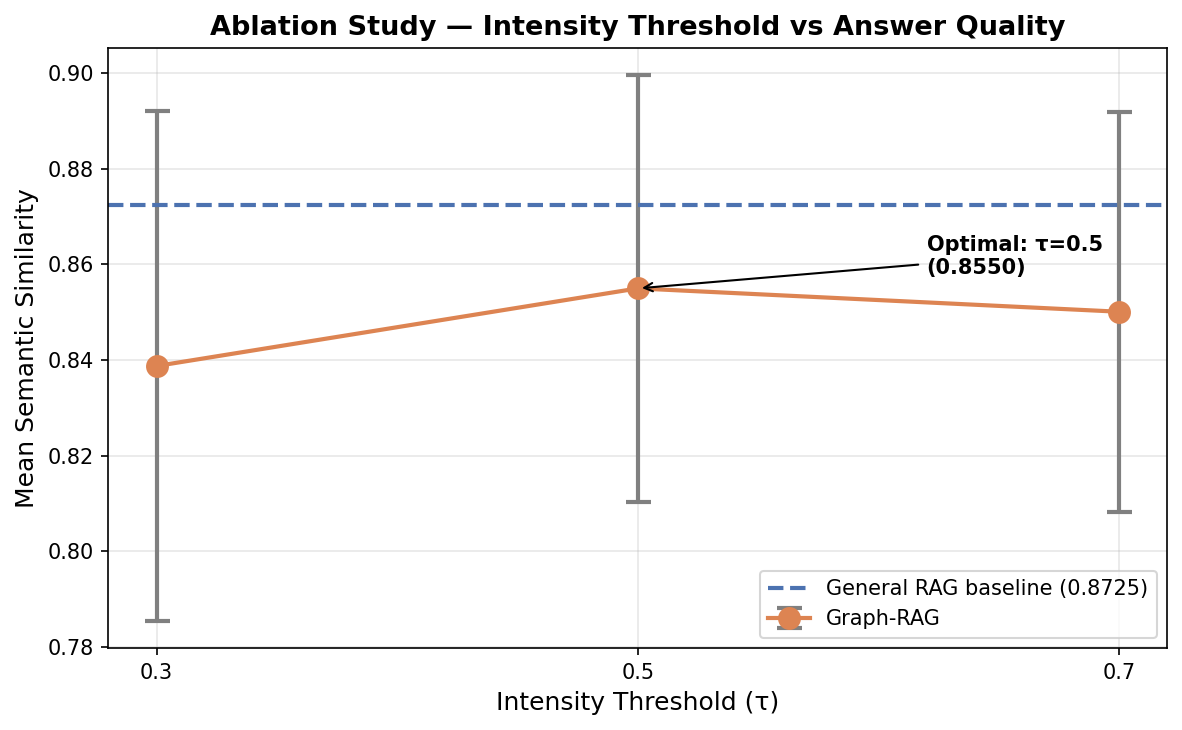}}
\caption{Inverted-U relationship between intensity threshold $\tau$ and semantic similarity. Optimal performance at $\tau = 0.5$; the General RAG baseline is shown as a dashed line.}
\label{fig:ablation}
\end{figure}

The results exhibit an inverted-U pattern, as illustrated in Fig.~\ref{fig:ablation}. At $\tau = 0.3$, the system traverses too many weak edges, introducing topically irrelevant evidence and degrading answer quality. At $\tau = 0.7$, the system is overly restrictive, missing moderately informative relationships. The optimal value of $\tau = 0.5$ balances breadth and precision, outperforming the production default of $\tau = 0.7$ by $+0.5\%$ in semantic similarity.

\subsection{Latency and Operational Efficiency}

Graph-RAG incurs a mean latency of $9.03$s per query compared to $7.37$s for General RAG—a $22.6\%$ overhead attributable to the Neo4j AuraDB network round-trip and the neighbor chunk re-ranking computation. However, Graph-RAG exhibits substantially lower latency variance ($\sigma = 0.96$s vs.\ $\sigma = 4.82$s), yielding more deterministic response times. The high variance in the General RAG pipeline is driven by fluctuations in Azure OpenAI API response latency, which dominates the end-to-end time when retrieval is instantaneous (FAISS in-memory).

In terms of context window utilization, Graph-RAG consumes $79\%$ more tokens on average ($1{,}711$ vs.\ $955$ tokens), resulting in a lower signal density ($1.689$ vs.\ $2.696$ relevant entities per $1{,}000$ tokens). This indicates that while Graph-RAG retrieves more total information, a portion of graph-traversed evidence is tangentially related rather than directly answering the query.

\subsection{Failure Case Analysis}

Of the 100 evaluation queries, Graph-RAG scored higher on 52, tied (within $\pm 0.01$ similarity) on 25, and was outperformed by General RAG on 23. Qualitative examination of the failure cases reveals two dominant patterns:

\begin{enumerate}
    \item \textbf{Single-entity factual queries:} Questions requiring precise information from one article, where additional cross-entity context dilutes the signal. Example: ``What was AAPL's Q3 revenue guidance?''
    \item \textbf{Low-connectivity entities:} Tickers with sparse or weak \texttt{INFLUENCES} edges in the graph, where the second hop produces minimal useful expansion.
\end{enumerate}

Conversely, Graph-RAG achieves its largest gains on queries involving supply chain dependencies, competitive dynamics, and sector-wide sentiment shifts—precisely the scenarios where relational context provides additive information unavailable through semantic similarity alone.

\section{Discussion}

\subsection{The Precision-Coverage Trade-off}

Our results demonstrate that Graph-RAG does not uniformly outperform vector-only retrieval. Instead, it shifts the operating point on a precision-coverage trade-off curve. For queries requiring cross-entity reasoning—which constitute the majority of real-world financial analyst questions—the 6.4\% improvement in entity recall provides meaningful value. For simple factual queries, the additional context introduces marginal noise without proportional benefit.

This finding aligns with the broader observation that RAG system design should be query-type-aware. A production system could route queries to the appropriate retrieval pipeline based on detected query complexity.

\subsection{Ablation as Architectural Guidance}

The inverted-U relationship between the intensity threshold and answer quality is a practical contribution for practitioners. It suggests that:

\begin{enumerate}
    \item Graph traversal filters should not be set at extreme values.
    \item The optimal threshold depends on graph density and edge quality—our finding of $\tau = 0.5$ is specific to our graph construction pipeline.
    \item Threshold tuning should be part of the standard Graph-RAG development workflow.
\end{enumerate}

\subsection{Limitations}

\textbf{LLM-generated ground truth.} Our evaluation set uses GPT-4o-mini-generated ideal answers, introducing a ceiling effect where both systems are evaluated against LLM output. While this limits absolute interpretability, relative comparisons between systems remain valid. Future work should incorporate human expert annotations.

\textbf{Scale.} Our evaluation uses 100 questions over 10 stocks. While sufficient for primary hypothesis testing (entity recall $p < 0.001$), subgroup analyses (e.g., per-stock breakdowns) lack statistical power. The ablation study on 20 questions should be interpreted as directional.

\textbf{Cloud latency.} Graph-RAG latency is inflated by Neo4j AuraDB network round-trips. An on-premises deployment would reduce this overhead, narrowing the latency gap.

\textbf{Single model.} All experiments use GPT-4o-mini. Results may vary with different LLMs, particularly those with larger context windows or stronger instruction-following capabilities.

\textbf{Domain specificity.} Our findings apply to technology sector financial news. Generalization to other domains (biomedical, legal) requires additional validation, though the architecture itself is domain-agnostic.

\section{Conclusion}

We presented a two-hop Graph-RAG architecture for financial sentiment analysis and provided a rigorous comparative evaluation against vector-only retrieval. Our key findings are:

\begin{enumerate}
    \item Graph-RAG achieves statistically significant improvement in entity recall (+6.4\%, $p < 0.001$), demonstrating its ability to surface structurally connected entities that vector retrieval misses.
    \item Graph-RAG improves Answer Relevancy by $+11.7\%$ overall, with the gain concentrating in relational queries ($+16.1\%$), indicating that graph-augmented context enables more topically focused multi-entity synthesis.
    \item This improvement comes at no measurable cost to answer quality ($\Delta = +0.001$ semantic similarity, $p = 0.281$, Cohen's $d = 0.078$) and a 22.6\% increase in latency.
    \item The intensity threshold exhibits an inverted-U relationship with answer quality, with $\tau = 0.5$ outperforming the default $\tau = 0.7$ by small margins.
    \item Graph-RAG's advantage concentrates in relational, multi-entity queries, while vector-only retrieval is sufficient for simple factual lookups.
\end{enumerate}

These results characterize a precision-for-coverage trade-off and provide actionable guidance for building retrieval systems in the financial domain. Future work will explore human evaluation, larger entity graphs, dynamic threshold selection, and multi-model generalization.

\bibliographystyle{IEEEtran}

\begin{thebibliography}{6}

\bibitem{lewis2020rag}
P.~Lewis \textit{et al.},
``Retrieval-augmented generation for knowledge-intensive NLP tasks,''
in \textit{Advances in Neural Information Processing Systems}, vol.~33, 2020.

\bibitem{barry2025graphrag}
M.~Barry \textit{et al.},
``GraphRAG: Leveraging graph-based efficiency to minimize hallucinations in LLM-driven RAG for finance data,''
Pre-print, 2025.

\bibitem{edge2024graphrag}
D.~Edge \textit{et al.},
``From local to global: A GraphRAG approach to query-focused summarization,''
Microsoft Research, \textit{arXiv preprint arXiv:2404.16130}, 2024.

\bibitem{es2023ragas}
S.~Es, J.~James, L.~Espinosa-Anke, and S.~Schockaert,
``RAGAS: Automated evaluation of retrieval augmented generation,''
\textit{arXiv preprint arXiv:2309.15217}, 2023.

\bibitem{chen2024knowledge}
Y.~Chen \textit{et al.},
``Knowledge-augmented financial market analysis and report generation,''
Tongji University / Ant Group, 2024.

\bibitem{nasiopoulos2025financial}
D.~K.~Nasiopoulos, K.~I.~Roumeliotis, D.~P.~Sakas, K.~Toudas, and P.~Reklitis,
``Financial sentiment analysis and classification: A comparative study of fine-tuned deep learning models,''
\textit{Int. J. Financial Stud.}, vol.~13, no.~2, p.~75, 2025.

\bibitem{zehra2021fkg}
S.~Zehra \textit{et al.},
``Financial knowledge graph based financial report query system,''
\textit{IEEE Access}, 2021.

\end{thebibliography}

\end{document}